%% file: acl2020.tex
\documentclass[11pt,a4paper]{article}
\usepackage[hyperref]{acl2020}
\usepackage{times}
\usepackage{latexsym}
\usepackage{subfig}
\usepackage{hyperref}

\usepackage[hang,flushmargin]{footmisc}  
\usepackage{amsfonts,amsmath,amssymb}
\usepackage[T1]{fontenc}  
\usepackage{bold-extra}   
\usepackage{microtype}    
\usepackage{graphicx}
\usepackage{multirow}
\usepackage{bm}           
\usepackage{soul}         

\aclfinalcopy 

\newcommand{\textsec}[1]{\textsection\ref{#1}}
\newcommand\LN{\linebreak\noindent}
\newcommand\CO{\texttt{CO}}
\newcommand\DS{\texttt{DS}}
\newcommand\EC{\texttt{EC}}
\newcommand\HP{\texttt{HP}}
\newcommand\OP{\texttt{OP}}
\newcommand\PG{\texttt{PG}}
\newcommand\SC{\texttt{SC}}
\newcommand\SW{\texttt{SW}}

\title{Noise Pollution in Hospital Readmission Prediction:\\Long Document Classification with Reinforcement Learning}

\author{
  Liyan Xu\textsuperscript{$\dagger$} \hfill 
  Julien Hogan\textsuperscript{$\ddagger$} \hfill 
  Rachel Patzer\textsuperscript{$\ddagger$} \hfill 
  Jinho D. Choi\textbf{\textsuperscript{$\dagger$}} \\
  \textsuperscript{$\dagger$}Department of Computer Science, \textsuperscript{$\ddagger$}Department of Surgery \\
  Emory University, Atlanta, US \\
  \texttt{\{liyan.xu,julien.hogan,rpatzer,jinho.choi\}@emory.edu}}

\date{}

\begin{document}
\maketitle

\input{tex/abstract}
\input{tex/introduction}
\input{tex/data}

\input{tex/related_work}
\input{tex/approach}

\input{tex/reinforcement-learning}

\input{tex/experiment}
\input{tex/analysis}
\input{tex/conclusion}
\input{tex/ack}

\bibliography{acl2020}
\bibliographystyle{acl_natbib}

\end{document}

%% file: tex/abstract.tex
\begin{abstract}
This paper presents a reinforcement learning approach to extract noise in long clinical documents for the task of readmission prediction after kidney transplant. 
We face the challenges of developing robust models on a small dataset where each document may consist of over 10K tokens with full of noise including tabular text and task-irrelevant sentences. 
We first experiment four types of encoders to empirically decide the best document representation, and then apply reinforcement learning to remove noisy text from the long documents, which models the noise extraction process as a sequential decision problem. 
Our results show that the old bag-of-words encoder outperforms deep learning-based encoders on this task, 
and reinforcement learning is able to improve upon baseline while pruning out 25\% text segments.
Our analysis depicts that reinforcement learning is able to identify both typical noisy tokens and task-specific noisy text.
\end{abstract}

%% file: tex/introduction.tex
\section{Introduction}

Prediction of hospital readmission has always been recognized as an important topic in surgery. 
Previous studies have shown that the post-discharge readmission takes up tremendous social resources, while at least a half of the cases are preventable \cite{Roy2015,Jones2016}. 
Clinical notes, as part of the patients' Electronic Health Records (EHRs), contain valuable information but are often too time-consuming for medical experts to manually evaluate. 
Thus, it is of significance to develop prediction models utilizing various sources of unstructured clinical documents.

The task addressed in this paper is to predict 30-day hospital readmission after kidney transplant, which we treat it as a long document classification problem without using specific domain knowledge. 
The data we use is the unstructured clinical documents of each patient up to the date of discharge. 
In particular, we face three types of challenges in this task. 
First, the document size can be very long; documents associated with these patients can have tens of thousands of tokens. 
Second, the dataset is relatively small with fewer than 2,000 patients available, as kidney transplant is a non-trivial medical surgery. Third, the documents are noisy, and there are many target-irrelevant sentences and tabular data in various text forms (Section~\ref{sec:data}).

The lengthy documents together with the small dataset impose a great challenge on representation learning. 
In this work, we experiment four types of encoders: bag-of-words (BoW), averaged word embedding, and two deep learning-based encoders that are ClinicalBERT \cite{DBLP:journals/corr/abs-1904-05342} and LSTM with weight-dropped regularization \cite{merity2018regularizing}. 
To overcome the long sequence issue, documents are split into multiple segments for both ClinicalBERT and LSTM  (Section~\ref{sec:repr}).

After we observe the best performed encoders, we further propose to combine reinforcement learning (RL) to automatically extract out task-specific noisy text from the long documents, as we observe that many text segments do not contain predictive information such that removing these noise can potentially improve the performance. We model the noise extraction process as a sequential decision problem, which also aligns with the fact that clinical documents are received in time-sequential order. At each step, a policy network with strong entropy regularization \cite{pmlr-v48-mniha16} decides whether to prune the current segment given the context, and the reward comes from a downstream classifier after all decisions have been made (Section~\ref{sec:rl}).

\begin{table*}[htbp!]
\centering\small
\begin{tabular}{c||r|r|l} 
\bf Type & \multicolumn{1}{|c}{\bf P} & \multicolumn{1}{|c}{\bf T} & \multicolumn{1}{|c}{\bf Description} \\
\hline\hline
\CO & 1,354 &  4,395.3 & Report for every outpatient consultation before transplantation \\ 
\hline
\DS &   514 &  1,296.7 & Summary at the time of discharge from every hospital admission happened before transplant \\
\hline
\EC & 1,110 &  1,073.6 & Results of echocardiography \\
\hline
\HP & 1,422 &  3,025.1 & Summary of the patient's medical history and clinical examination  \\
\hline
\OP & 1,472 &  4,224.8 & Report of surgical procedures \\
\hline
\PG & 1,415 & 13,723.4 & Medical note during hospitalization summarizing the patient's medical status each day \\
\hline
\SC & 2,033 &  1,189.2 & Report from the evaluation of each transplant candidate by the selection committee \\
\hline
\SW & 1,118 &  1,407.6 & Report from encounters with social workers \\ 
\end{tabular}
\caption{Statistics of our dataset with respect to different types of clinical notes. P: \# of patients, T: avg.\ \# of tokens, \CO: Consultations, \DS: Discharge Summary, \EC: Echocardiography, \HP: History and Physical, \OP: Operative, \PG: Progress, \SC: Selection Conference, \SW: Social Worker. The report for \SC\ is written by the committee that consists of surgeons, nephrologists, transplant coordinators, social workers, etc. at the end of the transplant evaluation. All 8 types follow the approximately 3:7 positive-negative class distribution.}
\label{table:stats}
\end{table*}

Empirical results show that the best performed encoder is BoW, and deep learning approaches suffer from severe overfitting under huge feature space in contrast of the limited training data. 
RL is experimented on this BoW encoder, and able to improve upon baseline while pruning out around 25\% text segments (Section~\ref{sec:experiment}). 
Further analysis shows that RL is able to identify traditional noisy tokens with few document frequencies (DF), as well as task-irrelevant tokens with high DF but of little information (Section~\ref{sec:analysis}).

%% file: tex/data.tex
\section{Data}
\label{sec:data}

This work is based on the Emory Kidney Transplant Dataset (EKTD) that contains structured chart data as well as unstructured clinical notes associated with 2,060 patients. 
The structured data comprises 80 features that are lab results before the discharge as well as the binary labels of whether each patient is readmitted within 30 days after kidney transplant or not where 30.7\% patients are labeled as positive.

The unstructured data includes 8 types of notes such that all patients have zero to many documents for each note type.
It is possible to develop a more accurate prediction model by co-training the structured and unstructured data; however, this work focuses on investigating the potentials of unstructured data only, which is more challenging.

\subsection{Preprocessing}
\label{sec:noisy}

As the clinical notes are collected through various sources of EMRs, many noisy documents exist in EKTD such that 515 documents are HTML pages and 303 of them are duplicates.
These documents are removed during preprocessing.
Moreover, most documents contain not only written text but also tabular data, because some EMR systems can only export entire documents in the table format.

\begin{table}[htp!]
\centering
\begin{tabular}{l} 
\hline
Lab Fishbone (BMP, CBC, CMP, Diff) and \\
critical labs - Last 24 hours 03/08/2013 12:45 \\
142(Na) 104(Cl) 70H(BUN) - 10.7L(Hgb) < \\
92(Glu)  6.5(WBC) 137L(Plt) 3.6(K) 26(CO2) \\
\hline
\end{tabular}
\caption{An example of tabular text in EKTD.}
\label{table:noiseexample}
\end{table}

\noindent While there are many tabular texts in the documents (e.g., lab results and prescription as in Table~\ref{table:noiseexample}), it is impractical to write rules to filter them out, as the exported formats are not consistent across EMRs.
Thus, any tokens containing digits or symbols, except for one-character tokens, are removed during preprocessing.
Although numbers may provide useful features, most quantitative measurements are already included in the structured data so that those features can be better extracted from the structured data if necessary.
The remaining tabular text contains headers and values that do not provide much helpful information and become another source of noise, which we handle by training a reinforcement learning model to identify them (Section~\ref{sec:rl}).

Table~\ref{table:stats} gives the statistics of each clinical note type after preprocessing.
The average number of tokens is measured by counting tokens in all documents from the same note type of each patient.
Given this preprocessed dataset, our task is to take all documents in each note type as a single input and predict whether or not the patient associated with those documents will be readmitted.


%% file: tex/related_work.tex
\section{Related Work}

\citet{Bonggun2019Multimodal} presented ensemble models utilizing both the structured and the unstructured data in EKTD, where separate logistic regression (LR) models are trained on the structured data and each type of notes respectively, and the final prediction of each patient is obtained by averaging predictions from each models. Since some patients may lack documents from certain note types, prediction on these note types are simply ignored in the averaging process. For the unstructured notes, concatenation of Term Frequency-Inverse Document Frequency (TF-IDF) and Latent Dirichlet Allocation (LDA) representation is fed into LR. However, we have found that the representation from LDA only contributes marginally, while LDA takes significantly more inferring time. Thus, we drop LDA and only use TF-IDF as our BoW encoder (Section~\ref{ssec:bow}).

Various deep learning models regarding text classification have been proposed in recent years. 
Pretrained language models like BERT have shown state-of-the-art performance on many NLP tasks \citep{devlin-etal-2019-bert}. 
ClinicalBERT is also introduced on the medical domain \citep{DBLP:journals/corr/abs-1904-05342}. 
However, deep learning approaches have two drawbacks on this particular dataset. First, deep learning requires large dataset to train, whereas most of our unstructured note types only have fewer than 2,000 samples. Second, these approaches are not designed for long documents, and difficult to keep long-term dependencies over thousands of tokens.

Reinforcement learning has been explored to combat data noise by previous work \citep{AAAI1816537,qin-etal-2018-robust} on the short text setting. A policy network makes decision left-to-right over tokens, and is jointly trained with another classifier. However, there is little investigation of using RL on the long text setting, as it still requires an effective encoder to give meaningful representation of long documents. Therefore, in our experiments, the first step is to select the best encoder, and then apply RL on the long document classification.

%% file: tex/approach.tex
\section{Document Representation}
\label{sec:repr}

\subsection{Bag-of-Words}
\label{ssec:bow}

For the baseline model, the bag-of-words representation with TF-IDF scores, excluding stopwords \citep{nothman-etal-2018-stop}, is fed into logistic regression (LR).
The objective is to minimize the negative log likelihood of the gold label $y_i$:
\begin{equation} \label{eq:bce}
    -\frac{1}{m} \sum_{i=1}^{m} [y_i \log{p(g_i)} + (1-y_i) \log{1 - p(g_i)}]
\end{equation}
where $g_i$ is the TF-IDF representation of $D_i$.
In addition, we experiment two common techniques in the encoder to reduce feature space: token stemming, and document frequency cutoff.

\subsection{Averaged Word Embedding}
\label{ssec:avg-emb}

Word embeddings generated by fastText are used to establish another baseline, that utilizes subwords to better represent unseen terms  \citep{bojanowski-etal-2017-enriching}. 
It is suitable for this task as unseen terms or misspellings frequently appear in these clinical notes.
The averaged word embedding is used to represent the input document consisting of multiple notes, which gets fed into LR with the same training objective.


\subsection{ClinicalBERT}
\label{sec:clinicalbert}

Following \citet{DBLP:journals/corr/abs-1904-05342}, the pretrained language BERT model \citep{devlin-etal-2019-bert} is first tuned on the MIMIC-III clinical note corpus \citep{mimiciii}, which has shown to provide better related word similarities in medical domains. 
Then, a dense layer is added on the \texttt{CLS} token of the last BERT layer. 
The entire parameters are fine-tuned to optimize the binary cross entropy loss, that is the same objective as Equation~\ref{eq:bce}.

Since BERT has a limit on the input length, the input document of each patient is split into multiple subsequences. Each subsequence is within the BERT length limit, and serves as an independent sample with the same label of the patient. The training data is therefore noisily inflated. The final probability of readmission is computed as follows:
\begin{equation} \label{eq:clinicalbert}
    p(y_i = 1 | g_i) = \frac{p_{\text{max}}^{n_i} + p_{\text{mean}}^{n_i} n_i/c}{1 + n_i/c}
\end{equation}
where $g_i$ is the BERT representation of patient $i$, $n_i$ is the corresponding number of subsequences, and $c$ is a hyperparameter to control the influence of $n_i$. $p_{\text{max}}^{n_i}$ and $p_{\text{mean}}^{n_i}$ are the max and mean probability across the subsequences, respectively.

The motivation behind balancing between the max and mean probability is that subsequences do not contain equal information. 
$p_{\text{max}}^{n_i}$ represents the best potential, while longer text should give more importance to $p_{\text{mean}}^{n_i}$, because $p_{\text{max}}^{n_i}$ is more easily affected by noise as the text length grows. Although Equation~\ref{eq:clinicalbert} seems intuitive, the use of pseudo labels on subsequences becomes another source of noise, especially when there are thousands of tokens; thus, the performance is uncertain.
Section~\ref{sec:dlexp} provides detailed empirical analysis for this model.

\begin{figure*}[htbp!]
\centering
\includegraphics[width=\textwidth]{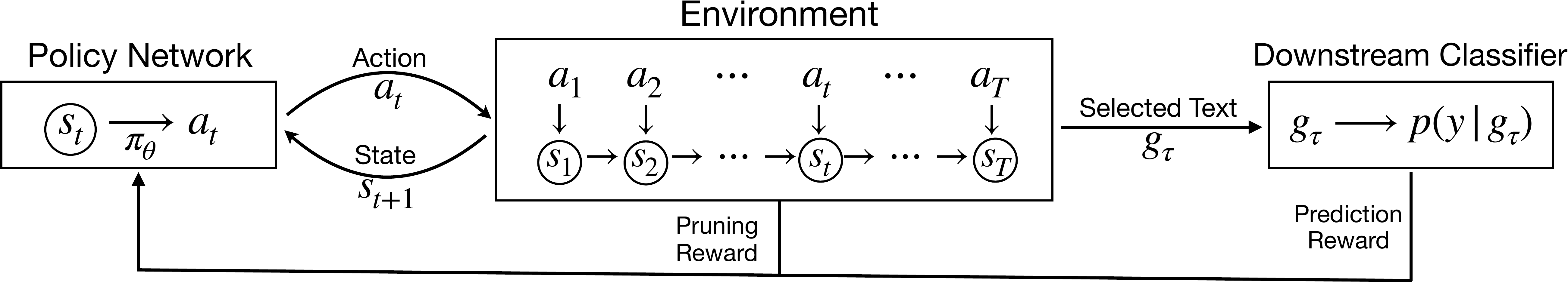}
\caption{Overview of our reinforcement learning approach. Rewards are calculated and sent back to the policy network after all actions $a_{1:T}$ have been sampled for the given episode.}
\label{fig:rl}
\end{figure*}

\subsection{Weight-dropped LSTM}
\label{sec:lstm}

We split documents of each patient into multiple short segments, and feed the segment representation to long short-term memory network (LSTM) at each time step:
\begin{equation}
    h_j \leftarrow \text{LSTM}(s_j, h_{j-1}; \theta)
\end{equation}
where $h_j$ is the hidden state at time step $j$, $s_j$ is the $j$th segment, and $\theta$ is the set of parameters.

\noindent Although segmentation of documents is still necessary, no pseudo labels are needed. We get the segment representation by averaging its token embedding from the last layer of BERT. The final hidden state at each step $j$ is the concatenated hidden states of a single-layer Bi-directional LSTM.
After we get the hidden state for each segment, a max-pooling operation is performed on $h_{1:n}$ over the time dimension to obtain a fixed-length vector, similar to \citet{kim-2014-convolutional,adhikari-etal-2019-rethinking}. A dense layer is immediately followed.

It is particularly important to strengthen regularization on this dataset with small sample size. Dropout \citep{JMLR:v15:srivastava14a} as a way of regularization has been shown effective in deep learning models, and \citet{merity2018regularizing} has successfully applied dropout-like technique in LSTM: the use of DropConnect \citep{10.5555/3042817.3043055} is applied on the four hidden-to-hidden matrices, preventing overfitting from occurring on the recurrent weights.

%% file: tex/reinforcement-learning.tex
\section{Reinforcement Learning}
\label{sec:rl}

Reinforcement learning is applied to the best performing encoder in Section~\ref{sec:repr} to prune noisy text, which can lead to comparable or even better performance, as many text segments in these clinical notes are found to be irrelevant to this task.
Figure~\ref{fig:rl} describes the overview of our reinforcement learning approach.
The pruning process is modeled as a sequential decision problem, for the fact that these notes are received in time-order. 
It consists of two separate components: a policy network, and a downstream classifier. 
To avoid having too many time steps, the policy is performed on the segment level instead of token level. For each patient, documents are split into short segments $g_{1:T} = \{g_1, g_2, \cdots, g_T\}$, and the policy network conducts a sequence of decisions $a_{1:T} = \{a_1, a_2, \cdots, a_T\}$ over segments. The downstream classifier is responsible for the reward, and the REINFORCE algorithm is used to train the policy \citep{10.1007/BF00992696}.

\paragraph{State} At each time step, the state $s_t$ is the concatenation of two parts: the representation of previously selected text, and the current segment representation $g_i$. The previously selected text serves as the context and provides a prior importance. Both parts are represented by an effective encoder, e.g. the best performing encoder from Section \ref{sec:repr}.

\paragraph{Action} The action space at each step is binary: \{\texttt{Keep}, \texttt{Prune}\}. If the action is \texttt{Keep}, the current segment is added to the selected text; otherwise, it is discarded. The final selected text for a patient is the concatenated segments selected by the policy.

\paragraph{Reward} The reward comes at the end when all actions are sampled for the entire sequence. The final selected text is fed to the downstream classifier, and negative log-likelihood of the gold label is used as the reward $R$. In addition, we also include a reward term $R_p$ to encourage pruning, as follows:
\begin{equation} \label{eq:reward}
    R_p = c \cdot \alpha \cdot [2\sigma(\frac{l}{\beta}) -1]
\end{equation}
where $c$ and $\beta$ are hyperparameters to control the scale of $R_p$, $l$ is the number of segments, $\alpha$ is the ratio of pruned segments $\left| \{a_k = \texttt{Prune}\} \right| / l$, $\sigma$ is the sigmoid function. The value of the term $2\sigma(\frac{l}{\beta}) -1$ falls into range $(0, 1)$. When $l$ is small, it downgrades the encouragement of pruning; when $l$ is large, it also gives an upper bound of $R_p$.
Additionally, we apply exponential decay on the reward. The final reward is $d^l R + R_p$. $d$ is the discount rate.

\begin{table*}[htbp!]
\centering
\begin{tabular}{l||c|c|c|c|c|c|c|c} 
\multicolumn{1}{c||}{\bf Encoder} & \bf\CO & \bf\DS & \bf\EC & \bf\HP & \bf\OP & \bf\PG & \bf\SC & \bf\SW \\ 
\hline\hline
Bag-of-Words (\textsec{ssec:bow})           &     58.6 &     62.1 &     52.0 &     58.9 &     51.8 &     61.2 & \bf 59.3 &     51.6 \\
$\:\:+$ Cutoff                              &     58.6 & \bf 62.3 &     52.8 &     59.0 &     51.9 &     61.3 & \bf 59.3 & \bf 51.9 \\
$\:\:+$ Stemming                            & \bf 58.9 &     61.8 & \bf 53.4 & \bf 59.4 &     51.9 & \bf 61.5 & \bf 59.3 &     51.6 \\
Averaged Embedding (\textsec{ssec:avg-emb}) &     56.3 &     53.7 &     52.4 &     54.0 & \bf 53.4 &     54.7 &     54.2 &     46.6 \\
\hline
ClinicalBERT (\textsec{sec:clinicalbert})   &     51.9 &     53.3 &     -    &     52.7 &     -    &     -    &     52.3 &     -    \\
Weight-dropped LSTM (\textsec{sec:lstm})    &     53.7 &     55.8 &     -    &     54.2 &     -    &     -    &     54.5 &     -    \\
\end{tabular}
\caption{The Area Under the Curve (AUC) scores achieved by different encoders on the 5-fold cross-validation. See the caption in Table~\ref{table:stats} for the descriptions of \CO, \DS, \EC, \HP, \OP, \PG, \SC, and \SW. For deep learning encoders, only four types are selected in experiments (Section \ref{sec:dlexp}).}
\label{table:encoders}
\end{table*}

\paragraph{Policy Network} The policy network maintains a stochastic policy $\pi(a_t|s_t;\theta)$:
\begin{equation}
    \pi(a_t|s_t;\theta) = \sigma(W s_t + b)
\end{equation}
where $\theta$ is the set of policy parameters $W$ and $b$, $a_t$ and $s_t$ are the action and state at the time step $t$ respectively. During training, an action is sampled at each step with the probability from the policy. After the sampling is performed over the entire sequence, the delayed reward is computed. During evaluation, the action is picked by $\text{argmax}_a \pi(a|s_t;\theta)$.

The training is guided by the REINFORCE algorithm \citep{10.1007/BF00992696}, which optimizes the policy to maximize the expected reward:
\begin{equation}
    J(\theta) = \mathbb{E}_{a_{1:T \sim \pi}} R_{a_{1:T}}
\end{equation}
and the gradient has the following form:
\begin{align} \label{eq:pg}
    \nabla_{\theta} J(\theta) &= \mathbb{E}_{\tau} \sum_{t=1}^T \nabla_{\theta} \log{\pi(a_t|s_t;\theta)} R_{\tau} \\
    &\approx \frac{1}{N} \sum_{i=1}^N \sum_{t=1}^T \nabla_{\theta} \log{\pi(a_{it}|s_{it};\theta)} R_{\tau_i}
\end{align}
where $\tau$ represents the sampled trajectory $\{a_1, a_2, \cdots, a_T\}$, $N$ is the number of sampled trajectories. $R_{\tau_i}$ here equals the delayed reward from the downstream classifier at the last step.

To encourage exploration and avoid local optima, we add the entropy regularization \citep{pmlr-v48-mniha16} on the policy loss:
\begin{equation}
    J_{reg}(\theta) = \frac{\lambda}{N} \sum_{i=1}^N \frac{1}{T_i} \nabla_{\theta} H(\pi(s_{it};\theta))
\end{equation}
where $H$ is the entropy, and $\lambda$ is the regularization strength, $T_i$ is the trajectory length. 

Finally, the downstream classifier and policy network are warm-started by separate training, and then jointly trained together.

%% file: tex/experiment.tex
\section{Experiments}
\label{sec:experiment}


Before experiments, we perform the preprocessing described in Section~\ref{sec:noisy}, and then randomly split patients in every note type by 5 folds to perform cross-validation as suggested by \citet{Bonggun2019Multimodal}.
To evaluate each fold $F_i$, 12.5\% of the training set, that is the combined data of the other 4 folds, are held out as the development set and the best configuration from this development set is used to decode $F_i$.
The same split is used across all experiments for fair comparison.
Following \citet{Bonggun2019Multimodal}, the averaged Area Under the Curve (AUC) across\LN these 5 folds is used as the evaluation metric.

\subsection{Baseline}

\paragraph{Bag-of-Words}
We first conduct experiments using the bag-of-words encoder (BoW; Section~\ref{ssec:bow}) to establish the baseline. 
Many experiments are performed on all note types using the vanilla TF-IDF, document frequency (DF) cutoff at 2 (removing all tokens whose DF $\leq 2$), and token stemming.\LN
For every experiment, the class weight is assigned inversely proportional to class frequencies, and the inverse of regularization strength $C$ is searched from $\{0.01, 0.1, 1, 10\}$, where the best results are achieved with $C = 1$ on the development set. 

Table~\ref{table:encoders} describes the cross-validation results on every note type.
The top AUC is $62.3\%$, which is within expectation given the difficulty of this task. 
Some note types are not as predictive as the others, such as Operative (\OP) and Social Worker (\SW), with the AUC under $52\%$. 
Most note types have the standard deviations in range $0.02$ to $0.03$.

In comparison to the previous work \citep{Bonggun2019Multimodal}, we achieve $0.671$ AUC combining both structured and unstructured data, despite without the use of LDA in our encoder.

\paragraph{Noise Observation} 
The DF cutoff coupled with token stemming significantly reduce feature space for the BoW model. 
As shown in Table \ref{table:vecsize}, the DF cutoff itself can achieve about 50\% reduction of the feature space. 
Furthermore, applying the DF cutoff\LN leads to slightly higher AUCs on most of the note types, despite almost a half of the tokens are removed from the vocabulary. 
This implies that there exists a large amount of noisy text that appears only in few documents, causing the models to be overfitted more easily. 
These results further verify our previous observation and strengthen the necessity to extract noise from these long documents using reinforcement learning (Section~\ref{ssec:exp-rl}).

\paragraph{Averaged Word Embedding}
For the averaged word embedding encoder (AWE; Section~\ref{ssec:avg-emb}), embeddings generated by FastText trained on the Common Crawl and the English Wikipedia with the 300 dimension is used.\footnote{\href{https://fasttext.cc/docs/en/crawl-vectors.html}{https://fasttext.cc/docs/en/crawl-vectors.html}}
AWE is outperformed by BoW on every note type except Operative (\OP; Table~\ref{table:encoders}). 
This empirical result implies that AWE over thousands of tokens is not so effective in generating the document representation so that the averaged embeddings are less discriminative than the sparse vectors generated by BoW for such long documents.

\begin{table}[htbp!]
\centering\resizebox{\columnwidth}{!}{
\begin{tabular}{c||r|r|r} 
\bf Type & \multicolumn{1}{c|}{\bf Vanilla} & \multicolumn{1}{c|}{\bf + Cutoff} & \multicolumn{1}{c}{\bf + Stemming} \\
\hline\hline
\CO & 28,213 & 15,022 (46.8) & 12,243 (56.6) \\
\DS & 11,029 &  6,117 (44.5) &  5,228 (52.6) \\
\HP & 20,245 & 11,276 (44.3) &  9,329 (53.9) \\
\SC & 19,050 &  9,873 (48.2) &  8,200 (57.0) \\
\end{tabular}}
\caption{The dimensions of the feature spaces used by each BoW model with respect to the four note types. The numbers in the parentheses indicate the percentage reduction from the vanilla model, respectively.}
\label{table:vecsize}
\end{table}

\subsection{Deep Learning-based Encoders}
\label{sec:dlexp}

For deep learning encoders, the four note types with good baseline performance ($\approx60\%$ AUC) and reasonable sequence length ($< 5000$) are selected to use in the following experiments, which are Consultations (\CO), Discharge Summary (\DS), History and Physical (\HP), and Selection Conference (\SC) (see Tables \ref{table:stats} and \ref{table:encoders}).

\paragraph{Segmentation} 
For both ClinicalBERT and the LSTM models, the input document is split into segments as described in Section~\ref{sec:clinicalbert}. 
For LSTM, we set the maximum segment length to be 128 for \CO\ and \HP, 64 for \DS\ and \SC, to balance between segment length and sequence length. The segment length for ClinicalBERT is set to 318 (approaching 500 after BERT tokenization) to avoid noise brought by too many pseudo labels.  More statistics about segmentation are summarized in Table \ref{table:seglen}.

For the ClinicalBERT, we use the PyTorch BERT implementation with the base configuration:\footnote{\href{https://github.com/huggingface/transformers}{https://github.com/huggingface/transformers}} 768 embedding dimensions and 12 transformer layers, and we load the weights provided by \citet{DBLP:journals/corr/abs-1904-05342} whose language model has been finetuned on large-scale clinical notes.\footnote{\href{https://github.com/kexinhuang12345/clinicalBERT}{https://github.com/kexinhuang12345/clinicalBERT}} We finetune the entire ClinicalBERT with batch size $4$, learning rate $2 \times 10^{-5}$, and weight decay rate $0.01$.

\noindent For the weight-dropped LSTM, we set the batch size to $64$, the learning rate to $10^{-3}$, the weight-drop rate to $0.5$, and search the hidden state dimension from $\{128, 256, 512\}$ on the development set. Early stop is used for both approaches.

\begin{table}[htbp!]
\centering
\begin{tabular}{c|r|r|r} 
\bf Type + Model & \multicolumn{1}{c|}{\bf SEN} & \multicolumn{1}{c|}{\bf SEQ} & \multicolumn{1}{c}{\bf INST} \\
\hline\hline
\CO\ + BERT & 318 & 14.8 & 11,376 \\
\CO\ + LSTM & 128 & 36.8 &    948 \\
\hline
\DS\ + BERT & 318 &  4.6 &  1,588 \\
\DS\ + LSTM &  64 & 22.5 &    371 \\
\hline
\HP\ + BERT & 318 & 10.1 &  8,364 \\
\HP\ + LSTM & 128 & 27.3 &    987 \\
\hline
\SC\ + BERT & 318 &  3.7 &  5,206 \\
\SC\ + LSTM &  64 & 25.4 &  1,422 \\
\end{tabular}
\caption{SEN: maximum segment length (number of tokens) allowed by the corresponding model, SEQ: average sequence length (number of segments), INST: average number of samples in the training set.}
\label{table:seglen}
\end{table}

\begin{figure}[htp]
\centering
\subfloat[Training loss]{%
  \includegraphics[width=0.5\columnwidth]{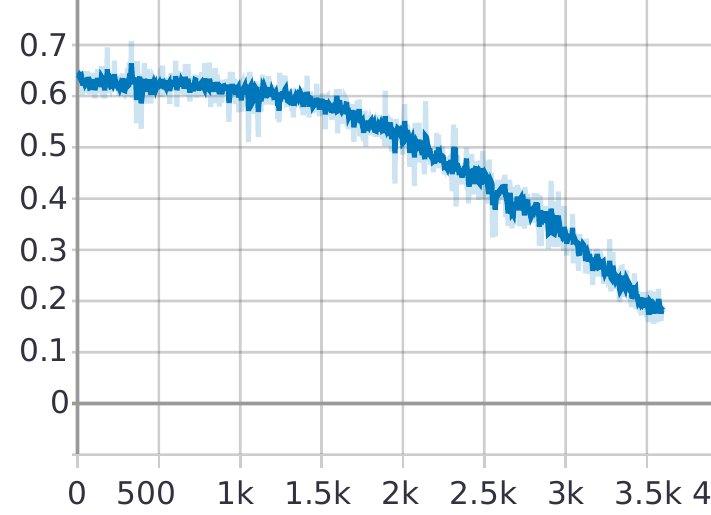}%
}
\subfloat[AUC on dev-set]{%
  \includegraphics[width=0.5\columnwidth]{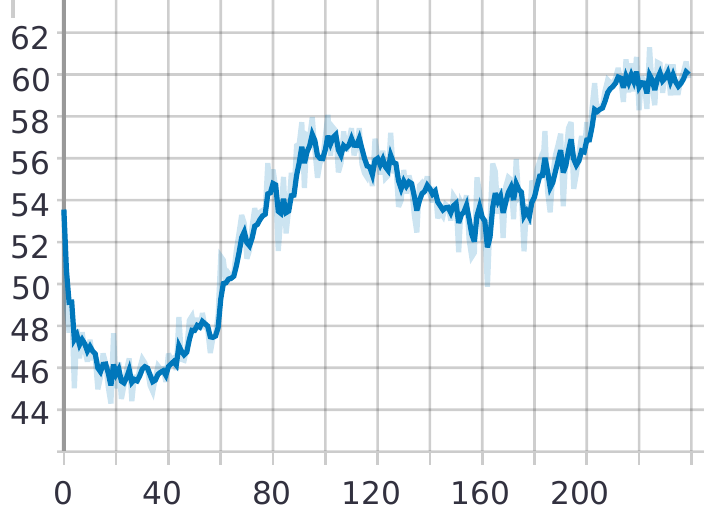}%
}
\caption{Training loss and AUC scores on the development set during the LSTM training on the \CO\ type. The AUC scores depict high variance while showing weak correlation to the training loss.}
\label{fig:dlplot}
\end{figure}

\paragraph{Result Analysis}
Table~\ref{table:encoders} shows the final results achieved by the ClinicalBERT and LSTM models.
The AUCs of both models experience a non-trivial drop from the baseline. 
After further investigation, the issue is that both models suffer from severe overfitting under the huge feature spaces, and struggle to learn generalized decision boundaries from this data. 
Figure \ref{fig:dlplot} shows an example of the weak correlation between the training loss and the AUC scores on the development set. 

\noindent As more steps are processed, the training loss gradually decreases to $0$. 
However, the model has high variance and it does not necessarily give better performance on the development set as the training loss drops. 
This issue is more apparent with ClinicalBERT on \CO\ because there are too many pseudo labels acting as noise, which makes it harder for the model to distinguish useful patterns from noise.

\subsection{Reinforcement Learning}
\label{ssec:exp-rl}

According to Table~\ref{table:encoders}, the BoW model achieves the best performance. Therefore, we decide to use TF-IDF to represent the long text of each patient, along with logistic regression as the classifier for reinforcement learning. 
Document segmentation is the same as LSTM (Table \ref{table:seglen}). During training, segments within each note are shuffled to reduce overfitting risks, and sequences with more than 36 segments are truncated.

The downstream classifier is warm-started by loading weights from the logistic regression model in the previous experiment. The policy network is then trained for 400 episodes while freezing the downstream classifier. After the warm start, both models are jointly trained. We set the number of sampling $N$ as $10$ episodes, learning rate $2 \times 10^{-4}$, and fix the scaling factor $\beta$ in Equations \ref{eq:reward} as $8$, and discount rate as $0.95$. Moreover, we search the reward coefficient $c$ in $\{0.02, 0.1, 0.4\}$, and entropy coefficient $\lambda$ in $\{2, 4, 6, 8\}$.

\begin{table}[ht]
\centering
\begin{tabular}{c||c|c|c|c} 
 & \bf\CO & \bf\DS & \bf\HP & \bf\SC \\
\hline\hline
Best & 58.9 & 62.3 & 59.4 & 59.3 \\
RL   & 59.8 & 62.4 & 60.6 & 60.2 \\
\hline
Pruning & 26\% &  5\% & 19\% & 23\% \\
\end{tabular}
\caption{The AUC scores and the pruning ratios of reinforcement learning (RL). Best: AUC scores from the best performing models in Table~\ref{table:encoders}.}
\label{table:rlresult}
\end{table}

\noindent The AUC scores and the pruning ratios (the number of pruned segments divided by the sequence length) are shown in Table~\ref{table:rlresult}. 
Our reinforcement learning approach outperforms the best performing models in Table~\ref{table:encoders}, achieving around 1\% higher AUC scores on three note types, \CO, \HP, and \SC, while pruning out up to 26\% of the input documents.

\paragraph{Tuning Analysis} We find that two hyperparameters are essential to the final success of reinforcement learning (RL).
The first is the reward discount rate $d$. The scale of the policy gradient $\nabla_{\theta} J(\theta)$ depends on the sequence length $T$, while the delayed reward $R_{\tau}$ is always on the same scale regardless of $T$. Therefore, different sequence length across episodes causes turbulence on the policy gradient, leading to unstable training. It is important to apply reward decay to stabilize the scale of $\nabla_{\theta} J(\theta)$.

\begin{figure}[htbp!]
\centering
\subfloat[Without Entropy Reg.]{%
  \includegraphics[width=0.5\columnwidth]{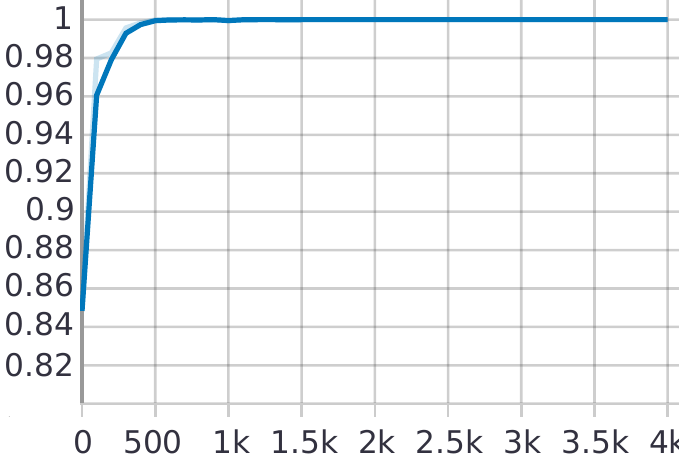}%
}
\subfloat[With Entropy Reg.]{%
  \includegraphics[width=0.5\columnwidth]{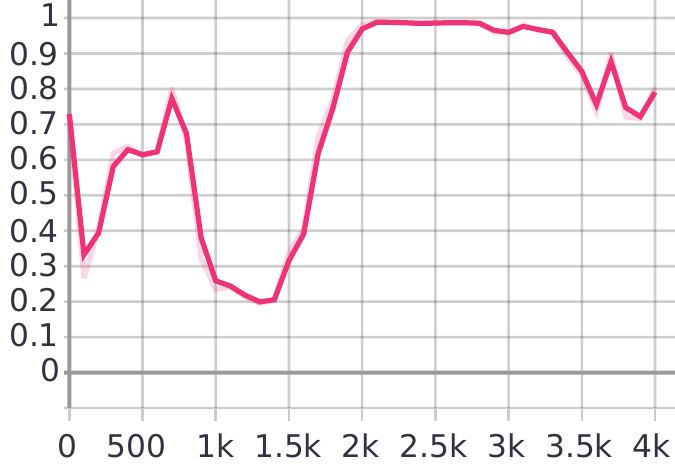}%
}
\caption{Retaining ratios on the development set of \SC\ while training the reinforcement learning model. Entropy regularization encourages more exploration.}
\label{fig:rltuning}
\vspace{-2ex}
\end{figure}

The second is the entropy regularization coefficient $\lambda$, which forces the model to add bias towards uncertainty. Without strong entropy regularization, the training is easy to fall into local optima in early stage, which is to keep all segments, as shown by Figure \ref{fig:rltuning}(a). $\lambda = 6$ gives the model descent incentive to explore aggressively, as shown by Figure \ref{fig:rltuning}(b), and finally leads to higher AUC.

%% file: tex/analysis.tex
\begin{table*}[ht]
\begin{center}
 \begin{tabular}{ p{0.95\textwidth}} 
 \hline\hline
\textbf{lab fishbone}  ( bmp , cbc , \textbf{cmp} , \textbf{diff} ) and critical labs - last hours ( not an official \textbf{lab report} . please see flowsheet ( or printed official \textbf{lab} reports ) for official \textbf{lab results} . ) ( na ) ( cl ) h ( \textbf{bun} ) - ( hgb ) ( glu ) ( wbc ) ( plt ) ( ) h ( \textbf{cr} ) ( hct ) na = not applicable a = abnormal ( ftn ) = \textbf{footnote} .\\
 \hline
\textbf{laboratory} studies : sodium , \textbf{potassium} , \textbf{chloride} , . , \textbf{bun} , \textbf{creatinine} , \textbf{glucose} . \textbf{total} bilirubin 1 , phos of , calcium , \textbf{ast} 9 , \textbf{alt} , alk phos . parathyroid hormone level . white blood cell count , \textbf{hemoglobin} , hematocrit , \textbf{platelets} . inr , ptt , and pt .\\
 \hline
 \textbf{methylprednisolone ivpb} : mg , \textbf{ivpb} , give in surgery , routine , / , infuse over : minute . \textbf{mycophenolate mofetil} : mg = 4 cap , po , capsule , once , now , / , stop date / , ml . documented medications documented accupril : mg , po , \textbf{qday} , 0 refill , substitution allowed .\\
 \hline\hline
\end{tabular}
\caption{Examples of pruned segments by the learned policy. Tokens that have feature importance lower than $-0.001$ (towards \textit{Prune} action) are marked bold.}
\label{table:pruned}
\end{center}
\end{table*}

\begin{table*}[ht]
\begin{center}
 \begin{tabular}{ p{0.95\textwidth}} 
 \hline\hline
the \textbf{social worker} met with this pleasant year old \textbf{caucasian male} on this date for kidney transplant \textbf{evaluation} . the patient was \textbf{alert} , oriented and easily \textbf{engaged} in conversation with the \textbf{social worker} today . he resides in atlanta with his spouse of years , who he describes as very \textbf{supportive} .\\
 \hline
 he \textbf{reports} occasional alcohol \textbf{drinks} per month but \textbf{denies} any illicit drug \textbf{use} . he has a \textbf{grade} education . he has been \textbf{married} for years . he is working full - time while on \textbf{peritoneal dialysis} as a business asset manager . he has \textbf{medicare} and an aarp \textbf{prescriptions} supplement . family history : mother \textbf{deceased} at age with complications of \textbf{obesity} , \textbf{high blood pressure} and \textbf{heart} disease .\\
 \hline\hline
\end{tabular}
\caption{Examples of kept segments by the learned policy. Tokens that have feature importance greater than $0.0005$ (towards \textit{Keep} action) are marked bold.}
\label{table:kept}
\end{center}
\vspace{-2ex}
\end{table*}

\section{Noise Analysis}
\label{sec:analysis}

To investigate the noise extracted by RL, we analyze the pruned segments on the validation sets of the Consultations type (\CO), and compare the results with other basic noise removal techniques.

\paragraph{Qualitative Analysis} Table \ref{table:pruned} demonstrates the potential of the learned policy to automatically identify noisy text from the long documents. The original notes of shown examples are tabular text with headers and values, mostly lab results and medical prescription. After the data cleaning step, the text becomes broken and does not make much sense for humans to evaluate. The learned policy can identify noisy segments by looking at the presence of headers such as ``lab fishbone'', ``lab report'', and certain medical terms that frequently appear in tabular reports such as ``chloride'', ``creatinine'', ``hemoglobin'', ``methylprednisolone'', etc. We find that many pruned segments have strong indicators of headers and specific medical terms, which appear mostly in tabular text rather than written notes.

Table \ref{table:kept} shows examples that are kept by the policy. Tokens that contribute towards \textit{Keep} action are words related with human and social life, such as ``social worker'', ``engaged'', ``drinks'', ``married'', ``medicare'', and terms related with health conditions, such as ``obesity'', ``heart'', ``high blood pressure''. These terms indeed appear mostly in written text rather than tabular data.

In addition, we also notice that the policy is able to remove certain duplicate segments. Medical professionals sometimes repeat certain description from previous notes to a new document, causing duplicate content. The policy learns to make use of the already selected context, and assigns negative coefficients to certain tokens. Duplicate segments are only selected once if the segment contains many tokens that have opposite feature importance in the context and segment vectors.

\begin{figure}[htp]
\centering
\includegraphics[width=0.85\columnwidth]{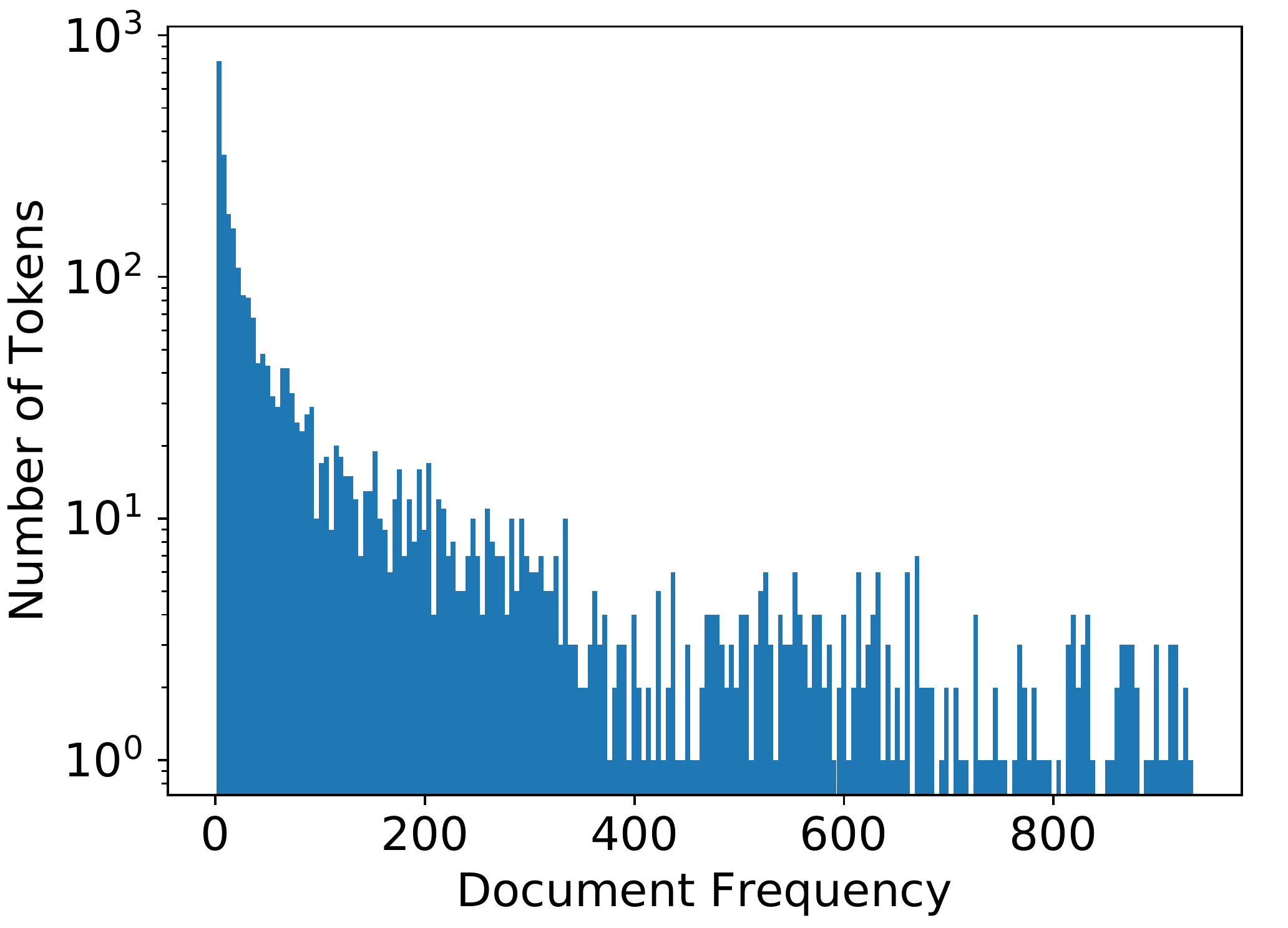}
\caption{Log scale distribution on document frequency of tokens with top negative feature importance.}
\label{fig:dfdist}
\vspace{-2ex}
\end{figure}

\paragraph{Quantitative Analysis} We examine tokens that are pruned by RL and compare with document frequency (DF) cutoff. We select 3000 unique tokens in the vocabulary that have the top negative feature importance (towards \textit{Prune} action) in the segment vector of \CO. Figure~\ref{fig:dfdist} shows the DF distribution of these tokens. 

\noindent We observe that the majority of those tokens have small DF values. It shows that the learned policy is able to identify certain tokens with small DF values as noise, which aligns with DF cutoff. Moreover, the distribution also shows a non-trivial amount of tokens with large DF values, demonstrating that RL can also identify task-specific noisy tokens that commonly appear in documents, which in this case are certain tokens in noisy tabular text.

Either RL or DF cutoff achieves higher AUC while reducing input features, proving that given the small sample size, the extracted text is more likely to cause overfit than being generalizable pattern, which also verifies our initial hypothesis.

%% file: tex/conclusion.tex
\section{Conclusion}

In this paper, we address the task of 30-day readmission prediction after kidney transplant, and propose to improve the performance by applying reinforcement learning with noise extraction capability. 
To overcome the challenge of long document representation with a small dataset, four different encoders are experimented. Empirical results show that bag-of-words is the most suitable encoder, surpassing overfitted deep learning models, and reinforcement learning is able to improve the performance, while being able to identify both traditional noisy tokens that appear in few documents, and task-specific noisy text that commonly appear.

%% file: tex/ack.tex
\section*{Acknowledgments}

We gratefully acknowledge the support of the National Institutes of Health grant R01MD011682, \textit{Reducing Disparities among Kidney Transplant Recipients}.
Any opinions, findings, and conclusions or recommendations expressed in this material are those of the authors and do not necessarily reflect the views of the National Institutes of Health.